\documentclass[letterpaper,10pt,conference]{IEEEtran}

\usepackage{copyright}

\usepackage[acronym,nonumberlist]{glossaries}
\usepackage{amsmath}
\usepackage{amssymb}
\newcommand{\R}{\mathbb{R}}
\usepackage[ruled,vlined]{algorithm2e}

\usepackage{glossary-mcols}
\setglossarystyle{mcolindex}

\newacronym{ood}{OoD}{Out-of-Distribution}
\newacronym{gan}{GAN}{Generative Adversarial Network}
\newacronym{vae}{VAE}{Variational Autoencoder}
\newacronym{ebm}{EBM}{Energy-Based Model}
\newacronym{cnn}{CNN}{convolutional neural network}
\newacronym{aspp}{ASPP}{Atrous Spatial Pyramid Pooling}
\newacronym{nn}{NN}{neural network}
\newacronym{dnn}{DNN}{deep neural network}
\newacronym{mlp}{MLP}{multilayer perceptron}
\newacronym{iou}{IoU}{intersection over union}
\newacronym{miou}{mIoU}{mean intersection over union}
\newacronym{bce}{BCE}{binary cross-entropy}
\newacronym{sota}{SOTA}{state-of-the-art}
\newacronym{sotif}{SOTIF}{``Safety of the Intended Functionality''}

\usepackage[binary-units]{siunitx}
\usepackage{textcomp}
\usepackage{eucal}
\usepackage{bbm}
\usepackage{graphicx}

\usepackage{cleveref}
\crefname{table}{Tab.}{Tabs.}
\crefname{figure}{Fig.}{Figs.}
\crefname{section}{Sec.}{Secs.}
\crefname{equation}{Eq.}{Eqs.}

\IEEEoverridecommandlockouts

\usepackage{cite}

\newcommand{\ourobj}{OoDCon}
\newcommand{\ouraug}{DomainMix}

\begin{document}

\title{\Large \bf Fool Me Once: Robust Selective Segmentation via\\Out-of-Distribution Detection with Contrastive Learning}
\author{David S. W. Williams, Matthew Gadd$^*$, Daniele De Martini$^*$, and Paul Newman\\
Oxford Robotics Institute, Dept. Engineering Science, University of Oxford, UK.\\\texttt{\{dw,mattgadd,daniele,pnewman\}@robots.ox.ac.uk}
\thanks{
$^*$ These authors contributed equally to this work.
}
}
\maketitle

\copyrightnotice

\begin{abstract}
In this work, we train a network to simultaneously perform segmentation and pixel-wise \gls{ood} detection, such that the segmentation of unknown regions of scenes can be rejected.
This is made possible by leveraging an \gls{ood} dataset with a novel contrastive objective and data augmentation scheme.
By combining data including unknown classes in the training data, a more robust feature representation can be learned with known classes represented distinctly from those unknown.
When presented with unknown classes or conditions, many current approaches for segmentation frequently exhibit high confidence in their inaccurate segmentations and cannot be trusted in many operational environments.
We validate our system on a real-world dataset of unusual driving scenes, and show that by selectively segmenting scenes based on what is predicted as \gls{ood}, we can increase the segmentation accuracy by an \acrshort{iou} of \num{0.2} with respect to alternative techniques.
\end{abstract}

\glsresetall

\section{Introduction}%
\label{sec:introduction}

A key task for a perception system is to segment the objects present in a scene; yet, machines deployed in environments with \emph{open-set conditions}~\cite{openset} are likely to eventually encounter objects which have never been seen before or which are well-known but have unusual appearance.
By implication, the data used to train the system is not fully representative of the data seen after deployment -- a deficiency often referred to as distributional shift. 
For a segmentation network to operate reliably under these conditions, it must assess its pixel-wise segmentation confidence, such that inaccurate segmentations of hitherto unseen objects and conditions can be rejected and perhaps flagged downstream in an autonomy stack.
This introspective capacity is a key requirement for the safe and reliable deployment of autonomous systems as presented in~\cite{limitspotentials,gadd2020evsav} and will  have an important role to play in satisfying the requirements in the international standards for functional safety in autonomous driving~\cite{ISO21448} -- this \gls{sotif} standard requires the minimisation of unknown unknowns by design.
We argue identifying likely performance via introspection measures is part of the answer.

\Glspl{nn}, which are nowadays the \gls{sota} representatives for such tasks, are typically trained for segmentation with cross-entropy loss and yield a categorical distribution over classes via softmax.
A reasonable entry-level introspective facility for \glspl{nn} is the identification of that which \emph{is not known} by assessing this distribution.
Unfortunately, segmentation networks are ``poorly calibrated'' as the entropy of the categorical distribution is a bad predictor of inaccurate segmentation.
Such overconfidence is often observed under data subject to distributional shift -- i.e. \gls{ood} -- or when ``attacked'' with adversarial examples.
This may occur because of: the extra degree of freedom in the softmax function~\cite{priornets}, insufficient regularisation leading to overfitting~\cite{on_the_calibration}, the ReLU activation functions~\cite{bad_relu}, and the cross-entropy objective~\cite{on_the_calibration,label_smoothing}. 


\begin{figure}[h]
\centering
\includegraphics[width=\columnwidth]{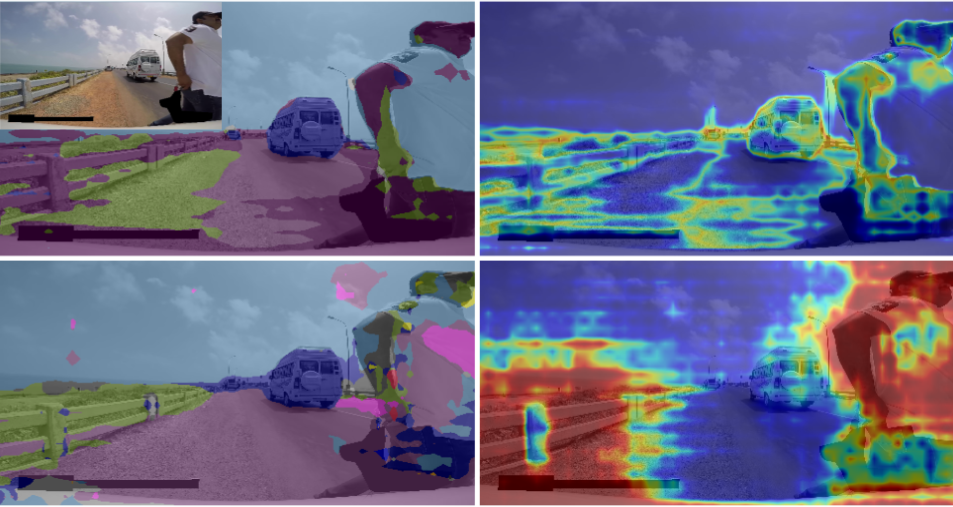}
\caption{
Demonstration of \gls{ood} detection being used to mitigate misclassification.
The RGB frame is shown inset top left.
Top row shows segmentation and uncertainty prediction from a baseline segmentation network from Hendrycks and Gimpel~\cite{baselineood}.
The baseline is highly confident in its classification of a pedestrian's shirt as sky.
Bottom row shows the segmentation and \gls{ood} prediction from our proposed system.
It detects the pedestrian as \gls{ood} having never seen one so close before, thereby mitigating misclassification.
}
\vspace{-.2cm}
\label{fig:uncertainty}
\end{figure}

In order to prevent false predictions from propagating out of the perception system and into autonomous decision-making, uncertainty estimation or \gls{ood} detection (also known as novelty or anomaly detection) can be employed.
While uncertainty estimation aims to directly predict the likelihood of error in the \gls{nn} output, \gls{ood} detection relates to predicting whether data belongs to the distribution of the training data.
As \glspl{nn} have a limited capacity to generalise, \gls{ood} data is very likely to result in error.

Uncertainty in \glspl{nn} is typically split into \textit{aleatoric} and \textit{epistemic}.
Aleatoric uncertainty is due to inherent ambiguity in the data -- e.g.~caused by environmental factors such as heavy rain, fog, darkness or motion and noise artefacts. 
Epistemic uncertainty, by contrast, relates to model parameter uncertainty, caused for instance by an imperfect training procedure or an insufficient quantity of suitable training data.
Often considered a subset of epistemic uncertainty~\cite{whatuncertainties}, \textit{distributional} uncertainty can be defined as uncertainty due to \gls{ood} data.

We believe that a requirement for a segmentation network operating in open-set conditions is the selective segmentation of a scene such that regions of high distributional or aleatoric uncertainty are detected and thereafter rejected.
We do not focus on epistemic uncertainty as it is reducible with an appropriate training procedure and large and diverse training datasets.
By contrast, distributional and aleatoric uncertainty are largely irreducible i.e. it is not possible to reduce the uncertainty in the segmentation of objects of unknown class or objects of known class in heavy darkness, thus demonstrating the need for rejection.



In this work, as demonstrated in~\cref{fig:uncertainty}, the effects of distributional uncertainty are mitigated by \emph{simultaneously} performing segmentation and pixel-wise \gls{ood} detection.
Our proposed contrastive loss leverages an \gls{ood} dataset and a novel data augmentation scheme, with the aim to learn a feature representation in which \gls{ood} and in-distribution data are reliably separable.
Unlike many techniques, our proposed technique also adds very little computational expense to the overall perception system, another important requirement for deployed mobile robotic systems.

This paper proceeds as follows.
\cref{sec:related} places the contribution in the available literature.
\cref{sec:system} describes our system.
\cref{sec:experiments} details our experimental setup and~\cref{sec:results} analyses the findings.
\cref{sec:concl} summarises the contribution and suggests further avenues for investigation.


\section{Related Work} \label{sec:related}
This section discusses our contribution in the context of the wealth of literature and treatments of deep uncertainty.
\subsection{Uncertainty Estimation}
On account of exact Bayesian inference being intractable, variational inference is used for epistemic uncertainty estimation by approximating the model parameter distribution, seen in Bayes-by-Backprop~\cite{blundell2015weight} and Monte Carlo dropout sampling~\cite{gal2016dropout}.
Alternatively,~\cite{deepensembles} performs epistemic uncertainty estimation with the parameter distribution defined over an ensemble of uniquely parameterised networks.


However, in both cases, multiple passes through a network are required.
This is prohibitively expensive to compute, particularly for a task such as segmentation with larger memory requirements and operating at a high frequency.
Therefore, the distillation of both dropout sampling and ensemble techniques into a single network has been proposed in~\cite{dropoutdistillation,ensembledistillation}.

Nevertheless,~\cite{bad_uncertainty} describes uncertainty estimation failure and worsening \gls{ood} detection performance as the distributional shift increases and the data becomes less familiar.

\subsection{Deep Generative Models}
Recently, deep generative models have shown the capacity to learn complex data distributions, allowing \gls{ood} samples to be detected.
This work includes \glspl{vae}~\cite{vaerecon2015}, \glspl{gan}~\cite{efficientGAN} and \glspl{ebm}~\cite{classifierEBM}.

However, scaling \gls{ood} detection with deep generative models to high resolution images is non-trivial.
Furthermore, as for uncertainty estimation, the robustness of deep generative models to distributional shift is concerning~\cite{bad_generative}.

\subsection{Contrastive Learning for OoD Detection}
Noise-contrastive estimation leverages explicit negative samples in order to learn better representations of data~\cite{noise_contrastive_estimation}.
Inspired by this, contrastive learning aims to learn a feature representation in which similar examples are close and dissimilar examples are farflung.
This is similar to deep generative modelling, except in its use of negatives instead of predefining the representation shape with probabilistic constraints (e.g. in \glspl{vae}, isotropic Gaussian latent distributions).
Additionally, the contrastive loss is computed on lower-dimensional intermediate feature maps as opposed to pixel-wise at the (high resolution) output (e.g. in \glspl{vae}, pixel-wise reconstruction).

A swathe of objective functions with differing similarity metrics and numbers of positive and negative examples have been presented.
In SimCLR~\cite{simclr}, an anchor image is augmented as a positive example and both are contrasted against many negative examples with dot product similarity.
In~\cite{supcon} it is shown that in a supervised setting, such a contrastive objective leads to a more robust representation, which can outperform standard cross-entropy.

Contrastive learning has had recent application to \gls{ood} detection~\cite{contrastive_ood1,contrastive_ood2}, albeit for image-wise classification and without direct use of \gls{ood} data during training.

\subsection{Novelty and Anomaly Detection}

Unsupervised techniques include using a proxy task to learn a useful feature representation~\cite{geometrictransformations}, learning a one-class classifier in an adversarial manner~\cite{adversarialOCC}, and an isolation forest model with a large \gls{cnn}~\cite{isolationforest}.

It is, however, unclear if image-wise techniques will scale to pixel-wise treatments.
Anomaly detection has been performed pixel-wise in video~\cite{videoanomaly1}.
This, however, requires a rarely available specialist labelled dataset to achieve good results.

Additionally, many of these techniques are so specialised for their task, that they would require a separate encoder from the segmentation task.
Unfortunately, this results in significantly larger memory requirements for the perception system.

Our approach solves both problems, leveraging a general \gls{ood} dataset and using the same encoder for both decoders.


\subsection{Softmax Calibration for Image-wise \gls{ood} Detection}

As a natural baseline,~\cite{baselineood} uses the maximum softmax score as a measure of confidence for detecting \gls{ood} images in an image-wise classification context.
\cite{odin} proposes ODIN, which uses temperature scaling and adversarial input preprocessing in order to improve the discriminative capacity of using softmax for \gls{ood} detection.
This line of research relies on the assumption that classification network's naturally represent in-distribution and \gls{ood} data in a reliably distinct way.
In this paper, we show that this is not the case by extending~\cite{baselineood} to segmentation and using it as a baseline, and show that improvements can be made by using \gls{ood} data.

\subsection{Data-Driven OoD Detection}
With the use of an \gls{ood} dataset, more robust feature detectors can be trained such that the learned representations of \gls{ood} data and in-distribution data are more separable.

In~\cite{outlierexposure}, it is shown how this improves both image-wise \gls{ood} detection and density estimation with generative models.
In~\cite{priornets}, a more calibrated image-wise classification network with a Dirichlet-distributed output is trained.
In~\cite{oodsegmentation}, pixel-wise \gls{ood} detection with a multi-head network is performed -- treating \gls{ood} detection as binary segmentation and using a simple \gls{bce} objective. 
In~\cite{kaistGANsoftmax}, the calibration of the softmax prediction of a classifier is improved with the use of generated \gls{ood} data.

Our work also leverages an \gls{ood} dataset in order to perform \gls{ood} detection.
However, it scales up the problem to pixel-wise \gls{ood} detection unlike many approaches for much simpler image-wise \gls{ood} detection.
In doing so, we propose a novel contrastive loss function coupled with a data augmentation scheme designed to train an encoder suited to both segmentation and robust pixel-wise \gls{ood} detection.
We show that these changes lead to an improvement in the network's ability to perform selective segmentation on a dataset that contains data at (and beyond) the limit of the training distribution.



\section{Proposed System Design}
\label{sec:system}

\begin{figure}[h]  
\centering
\includegraphics[width=\columnwidth]{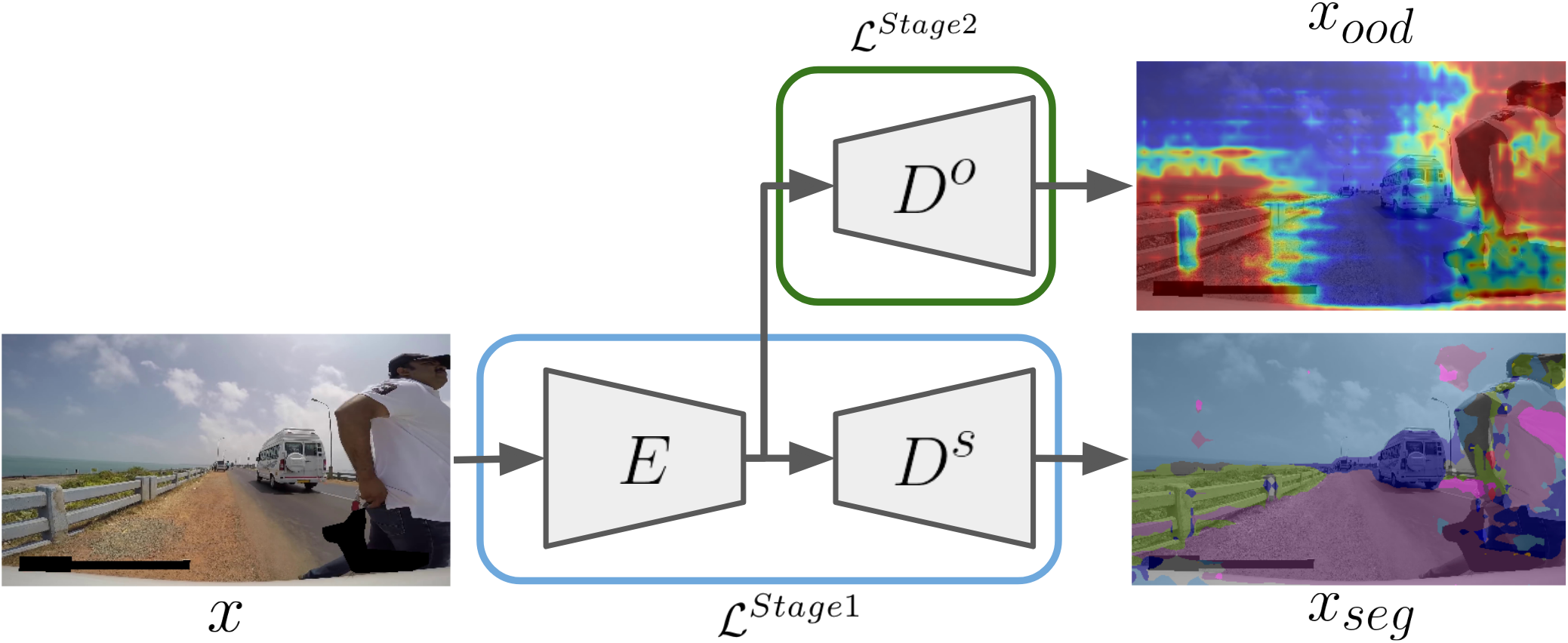}
\caption{
Overview of the proposed architecture with input-output examples same as shown in~\cref{fig:uncertainty}.
The encoder and segmentation decoder are jointly trained ($\CMcal{L}^{Stage1}$) to learn a feature representation suitable for both segmentation and pixel-wise \gls{ood} detection.
The \gls{ood} detection decoder is then trained ($\CMcal{L}^{Stage2}$) to output the final \gls{ood} prediction.\label{fig:networkdiagram}}
\vspace{-.2cm}
\end{figure}

Our segmentation network architecture is shown in~\cref{fig:networkdiagram}.

The encoder, $E: \R^{3 \times H \times W} \rightarrow \R^{C \times \hat{H} \times \hat{W}}$, maps an image to a high-dimensional feature map, where, $H,W$ and $\hat{H},\hat{W}$ are respectively the original and reduced spatial dimensions.
The segmentation decoder, $D^s: \R^{C \times \hat{H} \times \hat{W}} \rightarrow \R^{K \times H \times W}$, produces the final segmentation map, $x_{seg}$, where $K$ is the number of classes, $y$.
We view segmentation and \gls{ood} detection as distinct tasks and include a separate decoder for \gls{ood} detection, $D^o: \R^{\hat{H}\hat{W} \times C} \rightarrow \R^{1 \times H \times W}$, which takes a reshaped feature map and maps it to an \gls{ood} prediction, $x_{ood}$.

\subsection{Using OoD Data} \label{subsec:ooddata}
The in-distribution dataset for segmentation, $T^{ID}$, can be seen as the set, $T^{ID} \in \{i \in N| y_i \in K\}$, where $N$ is the set of all natural images, $y_i$ is the label for pixel $i$ and $K$ is the set of classes for the task.
Conversely, an \gls{ood} dataset for this task is then defined as $T^{OoD} \in \{i \in N| y_i \not\in K\}$.
The \gls{ood} dataset chosen for this task must therefore approximate the set of all natural images that does not contain the in-distribution segmentation classes -- implying significant scale and diversity.
Furthermore, pixel-wise \gls{ood} detection requires the separation of in-distribution and \gls{ood} \emph{within} images.

The use of a pixel-wise labelled dataset such as COCO~\cite{coco} is severely limiting.
We thus opt for a dataset uncurated for \gls{ood} detection, ImageNet~\cite{imagenet}, requiring no labels.
To combine the datasets, crops of images from each dataset are pasted into images of the other in the style of~\cite{cutmix}, followed by a novel data augmentation scheme in~\cref{subsec:data_aug}.

When using an uncurated \gls{ood} dataset, it is important to consider label noise -- i.e. that it contains pixels that are not truly \gls{ood}. 
This problem is ameliorated with a novel contrastive objective in~\cref{subsec:labelnoise}.



\subsection{OoD Detection via Contrastive Learning}
Extending a contrastive loss from image-wise classification to segmentation means the encoder yields a feature map with reduced spatial dimensions, $\R^{3 \times H \times W} \rightarrow \R^{C \times \hat{H} \times \hat{W}}$, instead of a single feature vector per image, $\R^{3 \times H \times W} \rightarrow \R^{C}$.
In order to perform pixel-wise \gls{ood} detection, the feature map is reshaped to produce $(\hat{H}*\hat{W})$ vectors of shape, $z \in \R^{C}$.

Our contrastive objective is designed to train an encoder that represents pixels from the training distribution with reliable separability from \gls{ood} data.
We start with the objective of~\cite{supcon} and define two classes: \textit{in-distribution} and \textit{\gls{ood}}.
Thus, we define $\CMcal{L}^{SupCon}$ as the total supervised contrastive loss over the batch $B, |B| = N$, as: 
\begin{equation} 
    \CMcal{L}^{SupCon} = \sum_{a\in B} \frac{1}{N_{y_a}-1} \: \CMcal{L}_a^{SupCon}
\end{equation}
where the loss $\CMcal{L}_a^{SupCon}$ related to anchor point $a\in B$ is:
\begin{equation}
    \CMcal{L}_a^{SupCon} = \sum\limits_{b \in B: b \neq a, y_b = y_a}
    -\log{\frac{\exp{(z^{\prime\top}_a z'_b/\tau)}}{\sum\limits_{c\in B: c \neq a}\exp{(z^{\prime\top}_a z'_c/\tau)}}}
\end{equation}

and $N_{\mathit{y}_a}$ is the number of feature vectors in the batch of the same class as anchor $a$.
The feature vector $z'$ is the mapping of $z$ by a projection head, $z' = P(z)$, as per~\cite{simclr}.


This objective pulls together examples from the same class in representation space and pushes away the other class.
However, we do not want to (and perhaps cannot) represent examples from the \gls{ood} class similarly as it contains a set of very diverse natural images -- dissimilar features are inevitable.
We therefore augment the objective function \emph{to not} enforce similarity between \gls{ood} feature representations and use them only as negatives.
This is essentially one-class contrastive learning with this class being a superset of the classes in the segmentation task.
In~\cref{eq:oodcon1}, we therefore only consider in-distribution feature vectors (with $y=0$) as anchor points.

Crucially, as described further in~\cref{subsec:labelnoise} below, we include a mask  -- $\mathbbm{1}_{LN}$ in~\cref{eq:oodcon2} -- to mitigate the effects of pixels \emph{incorrectly} defined as \gls{ood}:



\begin{equation}    \label{eq:oodcon1}
    \CMcal{L}^{OoDCon} = \sum_{a \in B: y_a = 0}^{N} \frac{1}{N_{y_a=0}-1} \: \CMcal{L}_a^{OoDCon}
\end{equation}

\begin{equation}    \label{eq:oodcon2}
    \CMcal{L}_a^{OoDCon} = \sum\limits_{b \in B: b \neq a, y_b = 0}
    -\log{\frac{\exp{(z_a^{\prime\top} z_b'/\tau)}}{\sum\limits_{c\in B: c \neq a}\mathbbm{1}_{LN}\exp{(z_a^{\prime\top} z_c'/\tau)}}}
\end{equation}

Intuitively, it may seem that pulling together feature representations of the known classes would ruin segmentation performance -- we would typically maximally separate them when using the cross-entropy loss for classification.
However,~\cite{softNNLoss2} shows that encouraging similarity between the known classes can be beneficial for the downstream task as it performs a kind of regularisation that promotes learning structure independent of class and aids in identifying \gls{ood} examples.

A key side effect of using the feature map from a segmentation encoder, $E: \R^{3 \times H \times W} \rightarrow \R^{C \times \hat{H} \times \hat{W}}$, in a contrastive loss is that much smaller batch sizes are used than, for example, in~\cite{supcon} for image classification. 
This is due to each image producing $(\hat{H}*\hat{W})$ times more vectors for contrasting than in the case of image classification.

\subsection{Masking Label Noise} \label{subsec:labelnoise}
As per~\cref{subsec:labelnoise}, this loss assumes significant label noise from the uncurated \gls{ood} dataset.
Thus, for the mask of~\cref{eq:oodcon2}, for each \gls{ood} feature vector, $i \in B: y_i = 1:$
\begin{equation}
        \mathbbm{1}_{LN} = \begin{cases} 0 & \text{if} \quad  \underset{j \in B: y_j = 0}{\max} (z_i^{\prime\top} z_j') > t_{R} \\
                                                                      1 & \text{otherwise}
                                                        \end{cases}
\end{equation}
Where $t_{R}$ is a threshold that rejects the proportion $R$ of the \gls{ood} features in $B$ and where $R$, the rejection ratio, is a hyperparameter set depending on the \gls{ood} dataset used.
In this work, $R=0.2$

This mask therefore ignores \gls{ood} feature vectors most similar to the in-distribution feature vectors being considered.
The metric used to rank the \gls{ood} features is the maximum similarity to the other in-distribution features in $B$.

For the original loss~\cite{supcon}, hard negatives and positives have large gradient contributions.
Without our mask, features from different classes which are nevertheless similar are considered hard negatives.
With the proposed masking, however, weight updates are not dominated by false hard negatives.

\subsection{Data Augmentation}\label{subsec:data_aug}
As per~\cref{subsec:ooddata}, the in-distribution and \gls{ood} datasets are combined within images.
Importantly, images are not just cropped into images from the other dataset (i.e. \gls{ood} into in-distribution), but also into images from the same dataset.
This prevents the automatic learning of the crop's origin and ensures that robust semantic features of the crop's content are learned.

Additionally, in order to alleviate characteristic large image gradients around crops, images are combined with a Gaussian blur kernel.
The mean hue and value of crops (in HSV) are also matched to the background region they replace, to make colour a less discriminative feature between crop and background.
These steps have the effect of making crops more similar to the background, leading to a more robust feature representation and more efficient training, as seen in~\cite{kaistGANsoftmax}, where \gls{ood} data is generated to be similar to in-distribution data but just outside of the distribution of known images. 

Finally, we randomly jitter the colour space of each image in brightness, contrast, saturation, and hue in order to reduce the average distance in colour space between images from both datasets.
We refer to this process as \ouraug{} data augmentation.
\begin{figure}[h]
\centering
\includegraphics[width=\columnwidth]{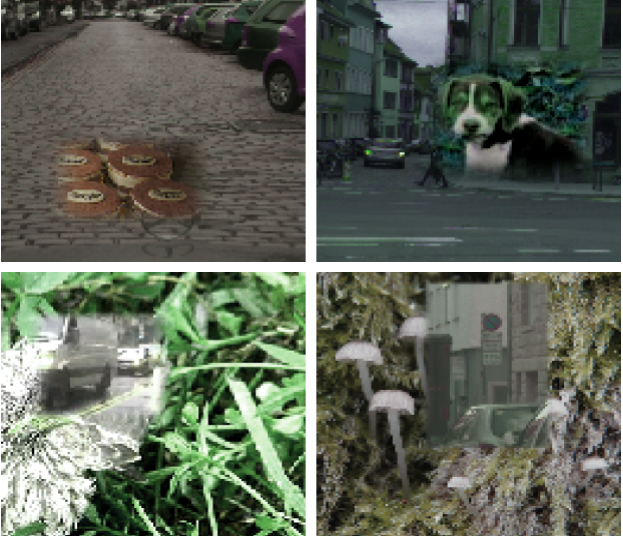}
\caption{Samples from the training dataset using \ouraug~data augmentation.
Images from the in-distribution driving dataset are combined with an \gls{ood} dataset, such that crops of \gls{ood} images are 
cut into in-distribution driving images (top) and vice versa (bottom).
Additionally, crops are blended in with a Gaussian kernel and colour space augmentations are performed to reduce the distance between the crop and the pixels it replaces.
Finally, random colour space augmentations are used to reduce learned dependence on colour as a discriminative feature. 
\label{fig:data_aug}}
\vspace{-.3cm}
\end{figure}

\subsection{Training Procedure}
Similarly to~\cite{supcon}, the training of the system is split into two stages, shown in~\cref{fig:networkdiagram}.
In both stages, the network is presented with augmented data (see~\cref{subsec:data_aug}).

In Stage 1, the encoder and segmentation decoder are trained by both the segmentation and contrastive objectives.
\begin{equation*}
    \CMcal{L}^{Stage 1} = \CMcal{L}^{seg} + \CMcal{L}^{OoDCon}
\end{equation*}
Here, $\CMcal{L}^{seg}$ is class-weighted cross-entropy, with weights calculated as in~\cite{enet}.
These objectives are mutually beneficial to the encoder.
The segmentation objective promotes learning features that allow the detection of the in-distribution classes, which in turn helps with identifying what is not a known class.
The contrastive objective promotes learning robust features of the in-distribution classes not present in the \gls{ood} data, such that their representations can be separated.
Additionally, by compressing the representations of the in-distribution classes, the contrastive objective provides useful regularisation~\cite{softNNLoss2}.

Stage 2 updates only the \gls{ood} decoder weights:
\begin{equation*}
    \CMcal{L}^{Stage 2} = \CMcal{L}^{BCE}
\end{equation*}
Here, $\CMcal{L}^{BCE}$ is the \gls{bce} loss function between the output of the \gls{ood} decoder, $x_{ood}$, and a binary label, $l_{ood}$, generated as part of the data augmentation process, where for a \gls{ood} pixel, $l_{ood} = 1$ else, $l_{ood} = 0$.

\section{Experimental Setup} \label{sec:experiments}

This section describes the experimental programme used to generate results which follow in~\cref{sec:results}.

\subsection{Data}
We use the Cityscapes~\cite{cityscapes} and Berkeley DeepDrive~\cite{bdd} datasets in order to train a \gls{cnn} to segment driving scenes.
These datasets both feature a ``void'' class which consists of all pixels that do not belong to any of the predefined classes.
For this reason, we consider the void class to be \gls{ood}.

With the \gls{ood} dataset, we approximate the set of all natural images that does not contain the known classes (see~\cref{subsec:ooddata}) using the ImageNet dataset~\cite{imagenet} due to its size and diversity of imagery.
It bears repeating (see~\cref{subsec:labelnoise}) that we treat this dataset as an uncurated set of random natural images and we do not make use of its labels --  no labelling is required for the \gls{ood} dataset.

In order to \emph{test} the trained networks, the WildDash dataset~\cite{wilddash} -- containing many unusual real-world driving scenes -- is held out.
We believe it is crucial to validate networks on real-world data that is at the limit of what is seen during training.
Indeed, detecting where misclassification is going to occur due to unfamiliarity is much more challenging than separating two distinct datasets combined by hand. 
We test on the 1000 most challenging images from this dataset, defined by those segmented least well by the baseline~\cite{baselineood}. 
During testing,~\cref{sec:results} shows that we reject predictions on image regions that are detected as \gls{ood}, reduce the misclassification risk, and improve segmentation performance.

Additionally, we investigate performance during a hand-crafted \gls{ood} detection task (as is more common in literature).
For this task, we leverage the pixel-wise labels in the COCO dataset~\cite{coco} in order to paste \gls{ood} images into the test splits of Cityscapes~\cite{cityscapes} and BerkeleyDeepDrive~\cite{bdd}.

\subsection{Performance Metrics}
On the \gls{ood} detection task, we use the best achieved average \gls{iou} across the generated dataset as a performance metric.
On the misclassification prediction task, we report segmentation performance in terms of the class \gls{miou} metric for a range of coverages.
Coverage is defined as the fraction of pixels that are not considered \gls{ood} and thus considered to be close enough to the training distribution to segment accurately.
Here, a certain coverage is reached by varying a threshold on the classification score from the \gls{ood} decoder.


\subsection{Network Architecture}
In each of our experiments, we use a DeepLabV3 segmentation network architecture~\cite{deeplabv3}.
It comprises a ResNet~\cite{he2016deep} encoder and an \gls{aspp} module as the segmentation decoder.
Specifically, we use a ResNet-18 encoder which produces a feature representation, $E:\R^{3 \times H \times W} \rightarrow \R^{512 \times \hat{H} \times \hat{W}}$, where $\hat{H}=H/8$, $\hat{W}=W/8$.

The \gls{ood} decoder, $D^o$, and projection head, $P$, are \glspl{mlp} with three and two hidden layers respectively.
Bilinear interpolation is employed in the \gls{ood} decoder to upsample to the same spatial dimensions as the input image. 
Both use batch normalisation and ReLU activation.
The projected feature vector, $z'$, has dimensionality, $z' \in \R^{64}$.
It is important to note that $z'$ is only computed during training -- not inference.


\section{Results} \label{sec:results}
The baseline is a segmentation network with no additional decoder, trained on only in-distribution data and uses $(1 - \operatorname{max~softmax~score})$ as a measure of uncertainty, as presented in~\cite{baselineood}.
In the two tests that follow, different objectives and data augmentation schemes used to enhance the baseline performance are investigated.
In terms of objectives, we compare the baseline and the proposed contrastive objective (\ourobj) to a KL divergence loss, which enforces a flat softmax response on \gls{ood} data at the end of the segmentation decoder, as seen in~\cite{kaistGANsoftmax}, and the \gls{bce} objective, performed on the \gls{ood} decoder~\cite{oodsegmentation}.
As for data augmentation, we include both our proposed method (c.f.~\cref{subsec:data_aug}, named \ouraug) and an adaptation of the Cutmix method~\cite{cutmix} to pixel-wise labelling.

\begin{table}[h!]
\renewcommand{\arraystretch}{1.5}
\centering
\caption{
Comparison of loss functions and data augmentation schemes for \gls{ood} detection task. We report detection performance with a measure of segmentation accuracy, \gls{iou}.
}
\label{tab:ood_detection_table}
\begin{tabular}{ccccc}
Objective        & Data Augmentation & IoU   \\ 
\hline
Baseline & - & 0.16        \\
BCE     & Cutmix & 0.26        \\
KL      & Cutmix & 0.27        \\
\ourobj  & Cutmix & 0.30        \\
BCE     & \ouraug & 0.41        \\
KL      & \ouraug & 0.35        \\
\ourobj  & \ouraug &\textbf{0.51}    \\
\vspace{-.6cm}
\end{tabular}
\end{table}

First, consider~\cref{tab:ood_detection_table}, which reports segmentation accuracy, \gls{iou}, on the \gls{ood} detection task, with each technique using a threshold corresponding to best achieved performance. 
It shows that the network trained with the contrastive loss and our proposed data augmentation scheme segments the inserted \gls{ood} data the most accurately.
The baseline performs poorly on the task compared to the other methods, suggesting a lack of inherent robustness in the feature representation for a network trained on in-distribution data alone, and that this can be improved with the use of \gls{ood} data. 

\begin{figure}[h!]
\centering
\includegraphics[width=\columnwidth]{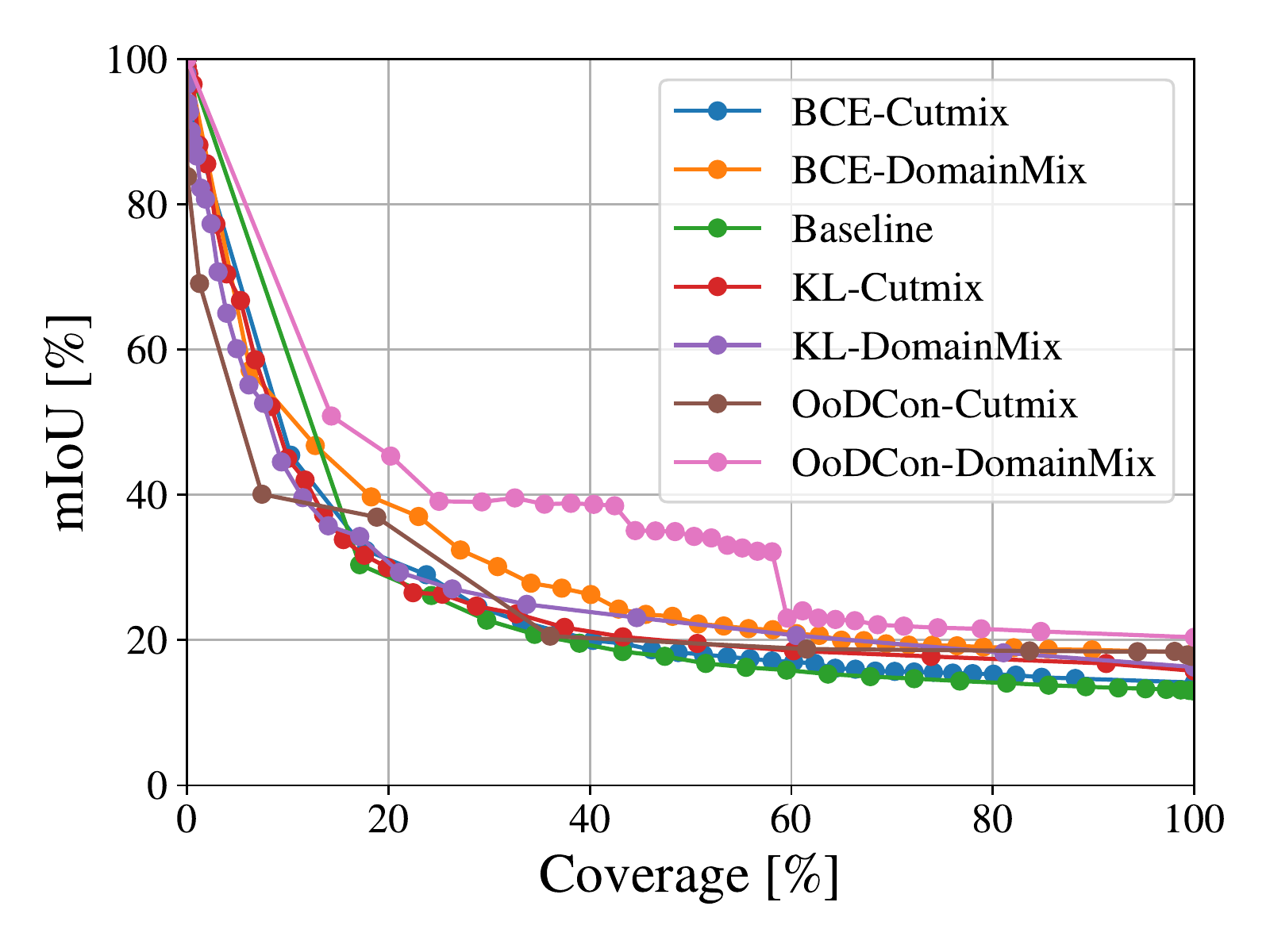}
\caption{
Segmentation accuracy (\gls{miou}) plotted as a function of the proportion of segmentations accepted (coverage).
Our proposed system corresponds to the objective ``OoDCon'' and data augmentation ``\ouraug'' row of~\cref{tab:ood_detection_table}.
As the threshold on classification score increases, the coverage decreases, more inaccurate segmentations are rejected and thus the remaining segmentations become more accurate, thereby increasing the mIoU.
When tested on challenging real-world data, our proposed network both segments the scenes most accurately and is better able to reject its erroneous segmentations, causing a significant increase in mIoU at a coverage of \SI{60}{\percent}.
\label{fig:results_plot}
}
\vspace{-.3cm}
\end{figure}

The results for the misclassification detection task are shown in~\cref{fig:results_plot}.
With no predictions rejected, the networks trained with our proposed data augmentation scheme can be seen to segment the challenging scenes best.
This suggests that \ouraug~promotes greater robustness to distributional shift in the learned feature representation due to the network's exposure to unknown classes and colour space shifts in training.

Upon rejecting the segmentations of \SI{40}{\percent} of pixels, a significant increase in segmentation accuracy is seen for the network trained with our proposed contrastive objective compared to the other techniques.
This means that the OoDCon objective has learned to represent unfamiliar pixels in a way that is more effectively separable from the in-distribution pixels, thus leading to the rejection of erroneous segmentations.

Indeed, the sharp knee seen at \SI{60}{\percent} coverage results in segmentation performance for our method which is only reachable at \SIrange{20}{30}{\percent} for the other methods - i.e. when rejecting \SIrange{70}{80}{\percent} of the \gls{nn} predictions.

In contrast to the \gls{ood} detection task, many of the other techniques do not perform better than the baseline.
This suggests that on the more challenging task of predicting misclassification on real-world data, only our proposed objective combined with our data augmentation generalises effectively from the \gls{ood} dataset to the real-world test dataset.  

\begin{figure}[h!]
\centering
\includegraphics[width=\columnwidth]{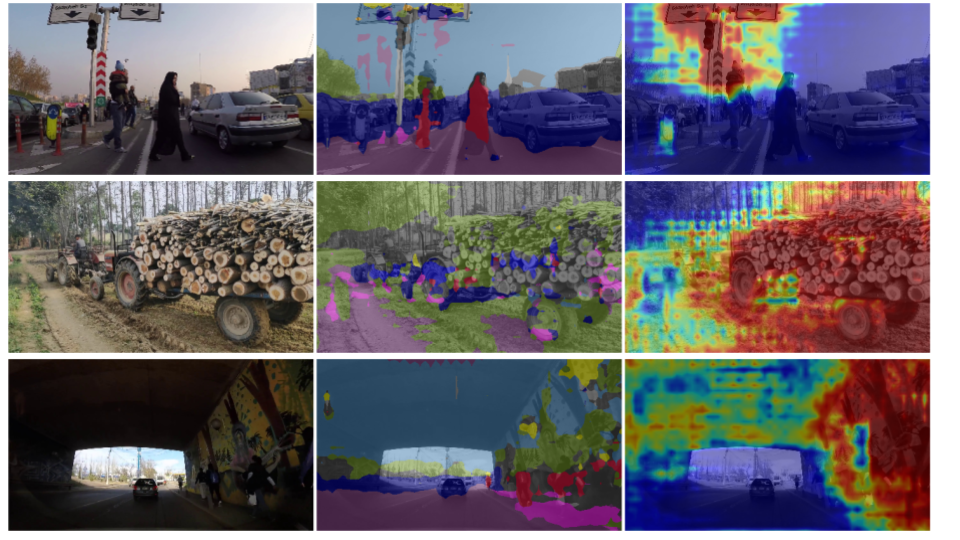}
\caption{
Qualitative results from our proposed technique on the WildDash test dataset~\cite{wilddash}, showing left to right: original image, segmentation and the \gls{ood} prediction.
Each image contains regions that are beyond the limit of the training distribution (top to bottom: uncommon signage, cart of chopped wood, and a heavily vandalised wall), resulting in poor, inconsistent segmentation.
By rejecting these regions based on our \gls{ood} prediction, inaccuracies can be prevented from propagating out of the perception system.
\label{fig:qual_results}
}
\vspace{-.5cm}
\end{figure}

Finally, consider the qualitative examples shown in~\cref{fig:qual_results}.
Regions of images from the WildDash dataset containing diverse unusualness (top to bottom: uncommon signage, cart of chopped wood, and a heavily vandalised wall) are detected as \gls{ood} having been segmented poorly. 

\Cref{fig:uncertainty} contains a comparison between the baseline and our proposed technique.
In this instance, our proposed segmentation network inaccurately segments a pedestrian, having never seen one so close, however the poor segmentation is mitigated by detecting the pedestrian as \gls{ood}. 
By contrast, the baseline confidently segments a pedestrian's shirt as sky in a manner unacceptable for a safety-critical task.

\section{Conclusion} \label{sec:concl}

In this work, we propose a segmentation network that has the capacity to mitigate misclassification by rejecting its predictions on \gls{ood} data, in a manner that adds no significant computational expense. 
This is achieved by using a novel contrastive objective along with an \gls{ood} dataset, presented with a novel data augmentation technique.
Upon testing with challenging real-world data, our proposed network is able to improve its performance significantly by selectively segmenting scenes when compared to proposed alternatives, thus proving its relative suitability for deployment on mobile robotic systems operating in open-set conditions.
Specifically, we show that by selectively segmenting scenes based on what is predicted as \gls{ood}, we can increase the segmentation accuracy by an \acrshort{iou} of \num{0.2} with respect to alternative techniques.
We expect that this contribution will advance the development of introspective performance assessment in scene understanding tasks.

In the future we will integrate this work in \emph{active} learning -- deploying the model, querying a human user upon surprising observation, and including this feedback in further training.

\section*{Acknowledgements}

This work was supported by the Assuring Autonomy International Programme, a partnership between Lloyd’s Register Foundation and the University of York, and UK EPSRC Programme Grant EP/M019918/1.

The authors are grateful to Oliver Bartlett for proofreading this manuscript.

\newpage
\bibliographystyle{IEEEtran}
\bibliography{biblio}

\end{document}